\documentclass[letterpaper, 10 pt, conference]{ieeeconf}  

\IEEEoverridecommandlockouts                              

\usepackage[dvipdfmx]{graphicx} 
\usepackage{times} 
\usepackage{amsmath} 
\usepackage{amssymb}  
\usepackage{amsfonts}
\usepackage{cite}
\usepackage{algorithm,algorithmic}
\usepackage{paralist}

\usepackage[table,xcdraw]{xcolor}
\usepackage{caption,subcaption}
\usepackage{comment}

\newcommand{\relmiddle}[1]{\mathrel{}\middle#1\mathrel{}}

\usepackage{amsmath}
\usepackage{amssymb}
\usepackage{accents} 

\newcommand{\eg}{\textit{e.g.,}~} %
\newcommand{\ie}{\textit{i.e.,}~} %

\newcommand{\bm}[1]{\textcolor{black}{#1}}
\newcommand{\oh}[1]{\textcolor{black}{#1}}

\DeclareMathOperator*{\argmax}{arg\,max}

\newcommand{\Z}       {\mathbf{Z}}             

\title{\LARGE \bf Bayesian Disturbance Injection: \\Robust Imitation Learning of Flexible Policies}

\author{Hanbit Oh$^{\dagger}$, Hikaru Sasaki, Brendan Michael and Takamitsu Matsubara$^{1}$
\thanks{$^{\dagger}$ Corresponding author} 
\thanks{$^{1}$ Authors are with the Division of Information Science, Graduate School of Science and Technology, Nara Institute of Science and Technology (NAIST),  Japan. (email: \tt\footnotesize{\{lastname.firstname.oe9, lastname.firstname.rw3, lastname.firstname, find.me.on.the.web\}@is.naist.jp})}
\thanks{© 2021 IEEE. Personal use of this material is permitted. Permission from IEEE must be obtained for all other uses, in any current or future media, including reprinting/republishing this material for advertising or promotional purposes, creating new collective works, for resale or redistribution to servers or lists, or reuse of any copyrighted component of this work in other works.}
\thanks{H. Oh, H. Sasaki, B. Michael and T. Matsubara, "Bayesian Disturbance Injection: Robust Imitation Learning of Flexible Policies," 2021 IEEE International Conference on Robotics and Automation (ICRA), 2021, pp. 8629-8635, doi: 10.1109/ICRA48506.2021.9561573.}
}

\begin{document}

\maketitle
\thispagestyle{empty}
\pagestyle{empty}

\begin{abstract}
Scenarios requiring humans to choose from multiple seemingly optimal actions are commonplace, however standard imitation learning often fails to capture this behavior. Instead, an over-reliance on replicating expert actions induces inflexible and unstable policies, leading to poor generalizability in an application. To address the problem, this paper presents the first imitation learning framework that incorporates Bayesian variational inference for learning flexible non-parametric multi-action policies, while simultaneously robustifying the policies against sources of error, by introducing and optimizing disturbances to create a richer demonstration dataset. This combinatorial approach forces the policy to adapt to challenging situations, enabling stable multi-action policies to be learned efficiently. The effectiveness of our proposed method is evaluated through simulations and real-robot experiments for a table-sweep task using the UR3 6-DOF robotic arm. Results show that, through improved flexibility and robustness, the learning performance and control safety are better than comparison methods.
\end{abstract}
\section{Introduction}
Imitation learning provides an attractive means for programming robots to perform complex high-level tasks, by enabling robots to learn skills via observation of an expert demonstrator \cite{coates2009apprenticeship,bojarski2016end,zhang2018deep,osa2018algorithmic}.
However, two significant limitations of this approach severely restrict applicability to real-world scenarios, these being: 
\begin{inparaenum}
\item learning policies from a human expert demonstrator is often very complex, as humans often choose between multiple optimal actions, and
\item learned policies can be vulnerable to compounding errors during online operation (\ie there is a lack of robustness).
\end{inparaenum}
In the former, uncertainties and probabilistic behavior of the human expert (\eg multiple optimal actions for a task), increases the complexity of learning policies, requiring \textit{flexible policy models}. In the latter case, compounding errors from environmental variations (\eg task starting position), may induce significant differences between the expert's training distribution and the applied policy. This drifting issue is commonly referred to as covariate shift \cite{ross2010efficient}, and requires \textit{robust policy models}.

To address both these limitations, this paper presents a novel Bayesian imitation learning framework, that learns a probabilistic policy model capable of being both flexible to variations in expert demonstrations, and robust to sources of error in policy application. This is referred to as Bayesian Disturbance Injection (BDI).
\begin{figure}[t]
    \centering
    \includegraphics[width=0.8\hsize]{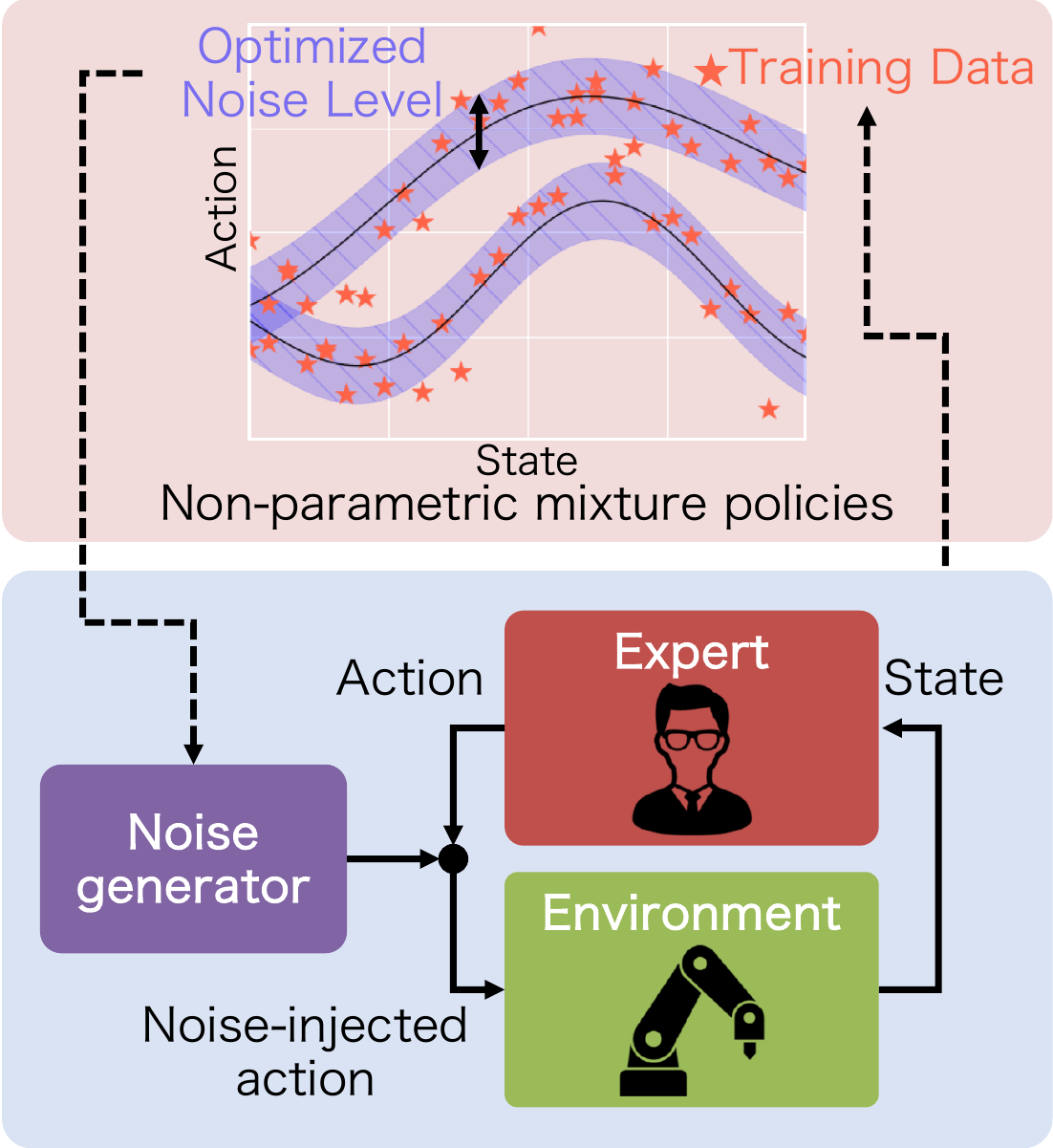}
    \caption{Overview of multi-modal policy learning with BDI.}
    \label{fig:overview}
\end{figure}

Specifically, this paper establishes robust multi-modal policy learning, with flexible regression models \cite{ross2013nonparametric} as probabilistic non-parametric mixture policies  (Fig.~\ref{fig:overview}, top). To induce robustness, noise is injected into the expert's actions (Fig.~\ref{fig:overview}, bottom) to generate a richer set of demonstrations. Inference of the policy, and optimization of injection noise is performed simultaneously by variational Bayesian learning, thereby minimising covariate shift between the expert demonstrations and learned policy.

To evaluate the effectiveness of the proposed framework, an implementation of Multi-modal Gaussian Process BDI (MGP-BDI) is derived, and experiments in learning probabilistic behavior from a table-sweeping task using the UR3 6-DOF robotic arm is performed. Results show improved flexibility and robustness, with increased learning performance and control safety relative to comparison methods.

\section{Related Work}
\subsection{Flexibility}
A key objective of imitation learning is to ensure that models can capture the variation and stochasticity inherent in human motion. \bm{Classical dynamical frameworks for learning trajectories from demonstrations include} \oh{Dynamic Movement Primitives (DMPs), which can generalize the learned trajectories to new situations (\ie goal location or speed). However, this generalization depends on the heuristic, thus unsuitable for learning state-dependent feedback policies \cite{schaal2005dmp,ijspeert2013dynamical,khansari2011learning}.}

\oh{Gaussian Mixture Regression (GMR) is an non-parametric, intuitive means to learn trajectories or policies from demonstrators in the state-action-space. In this, Gaussian Mixture Modelling (GMM) \cite{calinon2016tutorial} is used as a basis function to capture non-linearities during learning, and has been utilized in imitation learning that deals with human demonstrations \cite{kyrarini2019robot}. However, GMR requires to engineer features (\eg Gaussian initial conditions) by hand to deal with high-dimensional systems. \cite{huang2019kernelized}.}

As an alternative, Gaussian process regression (GPR) deals with implicit (high-dimensional) feature spaces with kernel functions. It thus can directly deal with high-dimensional observations without explicitly learning in this high-dimensional space \cite{rasmussen2003gaussian}. In particular, Overlapping Mixtures of Gaussian Processes (OMGP) \cite{lazaro2012overlapping} learns a multi-modal distribution by overlapping multiple GPs, and has been employed as a policy model with multiple optimal actions on flexible task learning of robotic policies \cite{sasaki2019multimodal}. To further reduce a priori tuning, Infinite Overlapping Mixtures of Gaussian Processes (IOMGP)  \cite{ross2013nonparametric} requires only an upper bound of the number of GPs to be estimated. As such, IOMGP is an intuitive means of learning complex multi-modal policies from unlabeled human demonstration data and is employed in this paper.

\subsection{Robustness}
While flexibility is key to capturing demonstrator motion, a major issue limiting application of learned policies is the problem of covariate shift. Specifically, environment variations (\eg manipulator starting position) induces differences between the policy distribution as learned by the manipulator and the actual task distribution during application.

A more general approach to minimizing the covariate shift in imitation learning, is Dataset Aggregation (DAgger) \cite{ross2011reduction}, whereby when the robot moves to a state not included in the training data, the expert teaches the optimal recovery. However, this approach has limited applicability in practice due to the risk of running a poorly learned robot and the tediousness to have human experts continue to teach the robot the optimal actions.

An intuitive approach to robustifying learned policies against sources of error, without needing to a priori specify task-relevant learning parameters, is to exploit phenomenon similar to persistence excitation \cite{sastry2011adaptive}. In this, noise is injected into the expert's demonstrated actions, and the recovery behavior of the expert is learned from this perturbation. In an imitation learning context, Disturbances for Augmenting Robot Trajectories (DART) \cite{laskey2017dart} exploits this phenomenon for learning a deterministic policy model with a single optimal action. Additionally, DART is well suited to creating a richer dataset, by  concurrently determining the optimal noise level to be injected into the demonstrated actions during policy learning.

However, the algorithm proposed to achieve DART \cite{laskey2017dart} has a major limitation. It assumes a uni-model deterministic policy, which is unsuitable for many real-world imitation learning tasks, that may consist of multiple optimal actions.

In contrast to that approach, this paper explores a novel method that can eliminate such a severe limitation by incorporating a prior distribution on policy then optimizing both the policy and the injection noise parameters simultaneously via variational Bayesian learning. As such, this novel imitation learning framework improves robustness while maintaining flexibility.
\section{Preliminaries}

\subsection{Imitation Learning from Expert's Demonstration}
The objective of imitation learning is to learn a control policy by imitating the action from the expert's demonstration data.
A dynamics model is denoted as Markovian with a state $\mathbf s_t \in \mathbb{R}^{Q}$, an action $a_t \in \mathbb{R}$ and a state transition distribution $p(\mathbf s_{t+1}\mid \mathbf s_t, a_t)$.
A policy $\pi(a_t\mid\mathbf s_t)$ decides an action from a state.
A trajectory $\tau=(\mathbf{s}_{0},{a}_{0},\mathbf{s}_{1},{a}_{1}\dots {a}_{T-1},\mathbf{s}_{T})$ which is a sequence of state-action pairs of $T$ steps.
The trajectory distribution is indicated as:
\begin{align}
    p(\tau\mid\pi)=p(\mathbf{s}_0)\prod_{t=0}^{T-1}\pi(a_t\mid\mathbf{s}_t)p(\mathbf{s}_{t+1}\mid\mathbf{s}_t,a_t).
\end{align}

A key aspect of imitation learning is to replicate the expert's behavior, and as such the function which computes the similarity of two policies using trajectories is defined as:
\begin{align}
    J(\pi,\pi^{\prime}\mid \tau) = -\sum_{t=0}^{T-1} \mathbb E_{\pi(a\mid \mathbf s_t),\pi^{\prime}(a'\mid \mathbf s_t)}\left[||a - a'||_{2}^{2}\right].
\end{align}
A learned policy $\pi^R$ is obtained by solving the following optimization problem using a trajectory collected by an expert's policy $\pi^*$:
\begin{align}\label{BC:policy}
    \pi^{R}= \argmax_\pi\mathbb{E}_{p(\tau\mid\pi^*)}\left[J(\pi,\pi^{*}\mid \tau)\right].
\end{align}

In imitation learning, a learned policy may suffer from the problem of error compounding, caused by covariate shift. This is defined as the distributional difference between the trajectories in data collection, and those in testing:
\begin{equation}
    \left|\mathbb{E}_{ p(\tau\mid\pi^*)}[J(\pi^{R},\pi^{*}\mid\tau)] -\mathbb{E}_{ p(\tau\mid\pi^R)}[J(\pi^{R},\pi^{*}\mid\tau)]\right|.
\end{equation}

\subsection{Robust Imitation Learning by Injecting Noise into Expert}
To learn policies that are robust to covariate shift, DART has previously been proposed \cite{laskey2017dart} for imitation learning problems. In this, expert demonstrations are injected with noise to produce a richer set of demonstrated actions. The level of injection noise is optimized iteratively to reduce covariate shift during data collection.

In this, it is assumed that the injection noise is sampled from a Gaussian distribution as $\epsilon_t\sim\mathcal N(0, \sigma^2_k)$, where $k$ is the number of optimizations. The injection noise $\epsilon_t$ is added to the expert's action $a^*_t$.

The noise injected expert's trajectory distribution is denoted as $p(\tau\mid\pi^*,\sigma^2_k)$ and the trajectory distribution with learned policy as $p(\tau\mid\pi^R_k)$.
To reduce covariate shift, DART introduces upper bound of covariate shift by Pinsker's inequality as:
\begin{align}\label{DART:covariate}
    &\left|\mathbb E_{p(\tau\mid\pi^*,\sigma^2_k)} \left[J(\pi^{R},\pi^{*}\mid\tau)\right]-\mathbb E_{p(\tau\mid\pi^R_k)} \left[J(\pi^{R},\pi^{*}\mid\tau)\right]\right|\nonumber\\
    &\leq T \sqrt{\frac{1}{2} \mathrm{KL}\left(p(\tau\mid\pi^R_k) \mid\mid p(\tau\mid\pi^*,\sigma^2_k)\right)},
\end{align}
where, $\mathrm{KL}(\cdot\mid\mid\cdot)$ is Kullback-Leibler divergence.
However, the upper bound is intractable since the learned policy's trajectory distribution $p(\tau\mid\pi^R_k)$ is unknown.
Therefore, DART solves the upper bound using the noise injected expert's trajectory distribution instead of the learned policy's trajectory distribution.
Optimal injection noise distribution is optimized as:
\begin{align}\label{DART:Noise}
    &\sigma_{k+1}^{2} =\argmax_{\sigma^{2}} \mathbb{E}_{p(\tau\mid\pi^{*},\sigma^2_{k})} \nonumber \\
    &~~~~~\left[\sum_{t=0}^{T-1} \mathbb E_{\pi^R_{k}(a_t'\mid\mathbf s_t)} [\log \mathcal{N}\left( a_t'\relmiddle| a_t, \sigma^{2}\right)]\right].
\end{align}
The optimized injection noise distribution can be interpreted as the likelihood of the learned policy.

The learned policy $\pi^R_k$ used in the optimization of $\sigma^2_{k+1}$ is obtained by following:
\begin{align}\label{DART:policy}
    \pi^R_k= \argmax_\pi\sum_{i=1}^{k-1}\mathbb{E}_{p(\tau\mid\pi^*,\sigma_i^2)}[J(\pi,\pi^{*}\mid\tau)].
\end{align}

\section{Proposed Method}
In this section, a novel Bayesian imitation learning framework is proposed (Fig. \ref{fig:overview}) to learn a multi-modal policy via expert demonstrations with noise injection.
\oh{Non-parametric mixture model is utilized as a policy prior and incorporates the injection noise distribution as a likelihood to a policy model.}
An imitation learning method is derived, which learns a multi-modal policy and an injection noise distribution, by variational Bayesian learning.
\oh{Note, IOMGP \cite{ross2013nonparametric} is employed as a policy prior in this paper. For simplicity but without loss of generality, this section focuses on one-dimensional action.}

\begin{figure}[t]
    \centering
    \includegraphics[width=0.7\hsize]{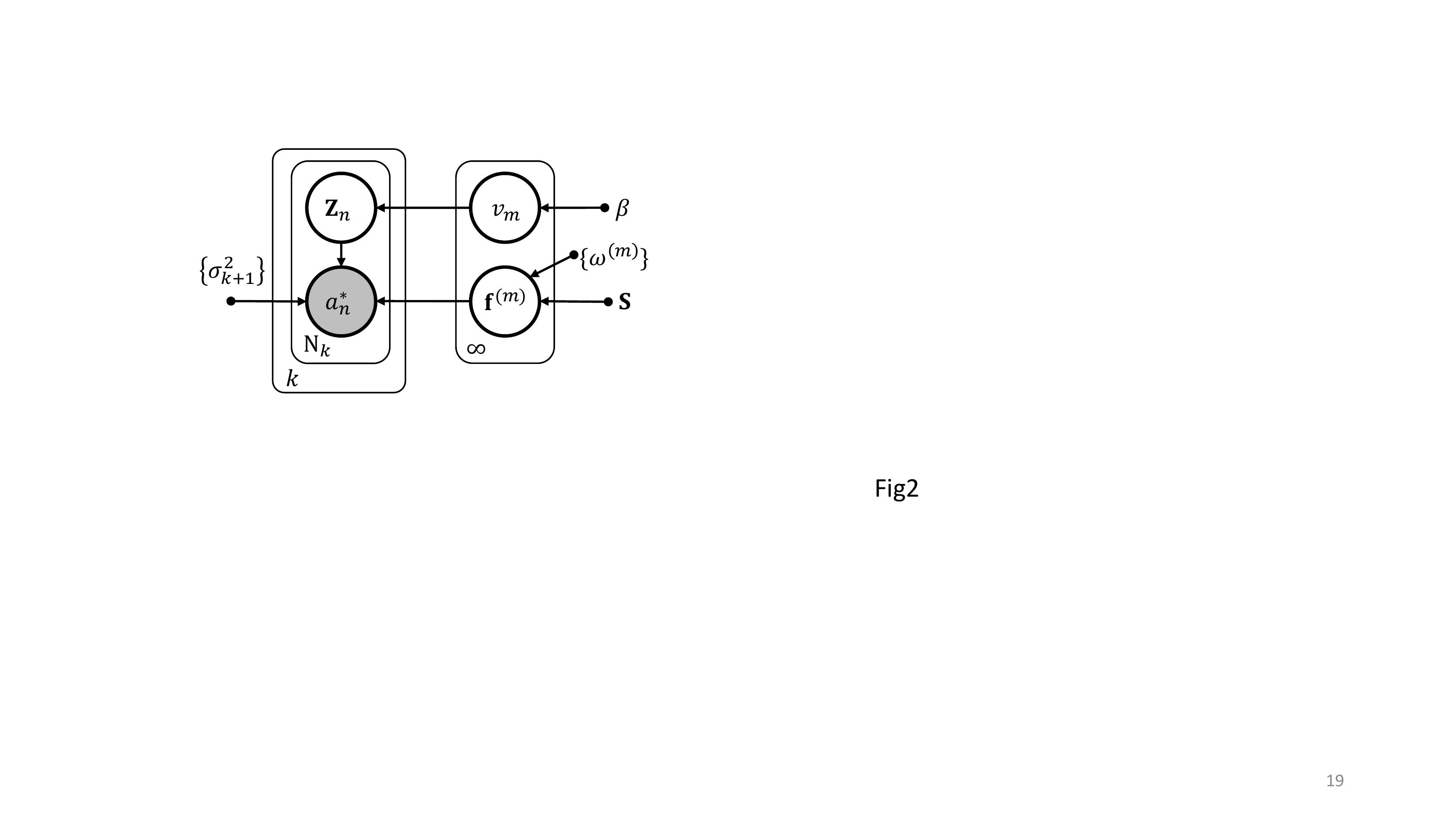}
    
    \caption{
    Graphical model of policy with injection noise parameter
    }
    \label{fig:grapicalmodel}
\end{figure}

\subsection{Policy Model}
\oh{To learn a multi-modal policy, the policy prior is considered as the product of infinite GPs, inspired by IOMGP. 
Fig.~\ref{fig:grapicalmodel} shows a policy model in which expert's actions $\mathbf{a^*}=[a^*_n]_{n=1}^{N}$ are estimated by $\mathbf{f}^{(m)}, \mathbf{Z},\{\sigma_{k+1}^2\}$ where $N = \sum_{j=1}^{k} N_j$ , $N_j$ is a size of the dataset that collected at $j$-th iteration.
The latent function $\mathbf{f}^{(m)}$ is the output of $m$-th GP given state $\mathbf{S}=[\mathbf{s}_n]_{n=1}^{N}$. 
To allocate $n$-th expert's action $a_n^*$ to $m$-th latent function $\mathbf{f}^{(m)}$, indicator matrix $\mathbf{Z} \in \mathbb{R}^{N\times \infty}$ is defined.
To estimate the optimal number of GPs, a random variable $v_m$ quantifies the uncertainty assigned to $\mathbf{f}^{(m)}$. 
In addition, the set of injection noise parameters, $\{\sigma_{k+1}^2\} = \{\sigma_{j+1}^2\}_{j=1}^{k}$, indicate the distribution of the expert's action $a_n^*$ in the distance from the latent function $\mathbf{f}^{(m)}$.}

The set of latent functions is denoted as $\{ {\mathbf{f}}^{(m)}\}=\{\mathbf{f}^{(m)}\}^{\infty}_{m=1}$ and a GP prior is given by :
\begin{equation}\label{prior_GP}
    p(\{\mathbf{f}^{(m)}\} \mid\mathbf{S}, \{\boldsymbol\omega^{(m)}\})=\prod_{m=1}^{\infty}  \mathcal{N}(\mathbf{f}^{(m)} \mid\mathbf{0},\mathbf{K}^{(m)}; \omega^{(m)}),
\end{equation}
\oh{where $\mathbf{K}^{(m)}=\mathrm{k}^{(m)}({\mathbf{S}}, {\mathbf{S}})$ is the $m$-th kernel gram matrix with the kernel function $\mathrm{k}^{(m)}(\cdot,\cdot)$ and a kernel hyperparameter $\omega^{(m)}$.}
Let $\{\boldsymbol{\omega}^{(m)}\} = \{\omega^{(m)}\}^{\infty}_{m=1}$ be the set of hyperparameters of infinite number of kernel functions.

Then the Stick Breaking Process (SBP) \cite{sethuraman1994constructive} is used as a prior of $\Z$, which can be interpreted as an infinite mixture model as follows:
\begin{align}\label{prior_z}
    p(\mathbf{Z} \mid \mathbf{v}) &= \prod_{n=1}^{N} \prod_{m=1}^{\infty}\left(v_{m} \prod_{j=1}^{m-1}\left(1-v_{j}\right)\right)^{\mathbf{Z}_{nm}}, \\
    p(\mathbf{v} \mid \beta)&=\prod_{m=1}^{\infty} \operatorname{Beta}\left(v_{m} \relmiddle| 1,\beta\right).
\end{align}
Note that the implementation of variational Bayesian learning cannot deal with infinite-dimensional vectors, so the $\infty$ component is replaced with a predefined upper bound of the mixtures $M$.
\oh{${{v}_m}$ is a random variable indicating the probability that the data corresponds to the $m$-th GP. Thus, it is possible to estimate the optimal number of GPs with a high probability of allocation starting from an infinite number of GPs. $\beta$ is a hyperparameter of SBP denoting the level of concentration of the data in the cluster.}

\oh{
The above policy model differs from the IOMGP model for regression \cite{ross2013nonparametric}; our model employs the set of injection noise parameters $\{{\sigma}_{k+1}^2\}$ because of the injection noise distribution is updated at each iteration.
The next injection noise optimized by (\ref{DART:Noise}) deals with the training data collected in the current iteration, while the policy inference (\ref{DART:policy}) deals with all the training data gathered up to the current iteration.}
Thus, the heteroscedastic Gaussian noise is defined as:
\begin{equation}
    \boldsymbol\Sigma=\mathrm{diag}\{\sigma_{i+1}\mathbf 1_{N_i}\}_{i=1}^k,
\end{equation}
showing the association between the injection noise parameters and the training data collected for each iteration. In this, $\mathbf 1_{N_i}$ is a vector whose size is $N_i$ and all components are one.
In addition, the value of the likelihood (\ref{DART:Noise}) does not change even if the mean and the input are replaced, the likelihood for the  $\{\mathbf{f}^{(m)}\}, \mathbf{Z}, \{{\sigma}_{k+1}^2\} $ is derived as :
\begin{align}\label{MGP-BDI:likelihood}
    &p({\mathbf{a}^*} \mid\{\mathbf{f}^{(m)}\}, \mathbf{Z}; \{{\sigma}_{k+1}^2\} ) \nonumber\\
    &=\prod_{n=1}^{N}\prod_{m=1}^{\infty} \mathcal{N}({a}_n^*\mid \mathbf{f}_n^{(m)},\boldsymbol\Sigma_{nn})^{\mathbf{Z}_{nm}}.
\end{align}

This formulation is described in a graphical model that defines the relationship between the variables as shown in Fig.~\ref{fig:grapicalmodel}, and the joint distribution of the model as :
\begin{align}\label{joint_distribution}
    &p({\mathbf{a}^*}, \mathbf{Z}, \mathbf{v},\{\mathbf{f}^{(m)}\}\mid\mathbf{S} ;\Omega ) \nonumber\\
    &=p({\mathbf{a}^*} \mid\{\mathbf{f}^{(m)}\}, \mathbf{Z}; \{{\sigma}_{k+1}^2\})\cdot \nonumber\\
    &~~~~~~~~p(\{\mathbf{f}^{(m)}\} \mid\mathbf{S}; \{\boldsymbol{\omega}^{(m)}\})
    p(\mathbf{Z} \mid \mathbf{v}) p(\mathbf{v} \mid \beta),
\end{align}
where $\Omega =(\{\boldsymbol{\omega}^{(m)}\},\{{\sigma}_{k+1}^2\},\beta)$ represents a set of hyperparameters.

\subsection{Optimization of Policies and Injection Noise via Variational Bayesian Learning}
Bayesian learning is a framework that estimates the posterior distributions of the policies and their predictive distributions for new input data rather than point estimates the policy parameters.
To obtain the posterior and predictive distributions, the marginal likelihood is calculated as :
\begin{align}\label{marginal_likelihood}
    &p({\mathbf{a}^*}\mid\mathbf{S};\Omega )\nonumber\\
    &= \iiint p({\mathbf{a}^*}, \mathbf{Z}, \mathbf{v},\{\mathbf{f}^{(m)}\}\mid\mathbf{S} ;\Omega )
    \mathrm d \mathbf{Z} \mathrm d \mathbf{v} \mathrm d\{\mathbf{f}^{(m)}\}.
\end{align}
However, it is intractable to calculate the log marginal likelihood of (\ref{marginal_likelihood}) analytically.
Therefore, the variational lower bound is derived as the objective function of variational learning. 
The true posterior distribution is approximated by the variational posterior distribution, which maximizes the variational lower bound.
\oh{As a common fashion of variational inference, the variational posterior distribution is assumed to be factorized among all latent variables (known as the \textit{mean-field approximation} \cite{parisi1988statistical}) as follows:}
\begin{equation}
    q(\mathbf{Z, v, \{{f}^{(m)}\}}) = q(\mathbf{Z}) \prod_{m=1}^\infty q(v_m) q(\mathbf {f}^{(m)}).
\end{equation}
The variational lower bound $\mathcal{L}(q,\Omega)$ is derived from this assumption by applying the Jensen inequality to the log marginal likelihood, as:
\begin{align}\label{variational_lower_bound}
    &\log {
    p({\mathbf{a}^*}\mid\mathbf{S};\Omega )
    }\nonumber\\
   &\geq  \iiint q\log 
   \frac{p({\mathbf{a}^*}, \mathbf{Z}, \mathbf{v},\{\mathbf{f}^{(m)}\}\mid\mathbf{S};\Omega)}{q}
   \mathrm d \mathbf{Z} \mathrm d \mathbf{v}\mathrm d\{\mathbf{f}^{(m)}\}\nonumber\\
   &=\mathcal{L}(q,\Omega).
\end{align}
In addition, the optimization formulation is derived using the \textit{Expectation-Maximization} (EM) algorithm.
The variational posterior distribution $q$ is optimized with a fixed hyperparameter $\Omega$ in E-step, and the hyperparameter $\Omega$ is optimized with a fixed variational posterior distribution $q$ in M-step with:
\begin{align}\label{MGP:objective_function}
    (\hat{q},\hat{\Omega}) = \argmax_{q,\Omega}
    \mathcal{L}(q,\Omega).
\end{align}
See the Section\ref{Appendix}-{A,B} for details of $q$ update laws and lower bound of marginal likelihood.
And a summary of the proposed method is shown in Algorithm~\ref{algorithm}.
\begin{algorithm}[tb]
\caption{MGP-BDI}
\label{algorithm}
\begin{algorithmic}[1]
\small
\renewcommand{\algorithmicrequire}{\textbf{Input:}}
\renewcommand{\algorithmicensure}{\textbf{Output:}}
\REQUIRE $M,K$
\ENSURE {$q,\hat{\Omega}$}
\FOR {$k = 1$ to $K$}
    \STATE Get dataset through the noise injected expert:\\ $\{a^*_t,\mathbf{s}_t\}_{t=1}^{N_k} \sim p(\tau\mid\pi^*,\sigma_k^2)$
    \STATE Aggregate datasets :$\mathcal{D} \gets \mathcal{D}\cup \{a^*_t,\mathbf{s}_t\}_{t=1}^{N_k}$ 
\WHILE {$ \mathcal{L}(q,\Omega)$ is not converged}
    \WHILE{$ \mathcal{L}(q,\Omega)$ is not converged}
    \STATE Update $q(\mathbf{f}^{(m)})$, $q(\mathbf{Z})$,and $q(v_m)$ alternately
    \ENDWHILE
    \STATE{Optimize $\Omega$ with fixed $q$:} $\hat{\Omega}\gets \argmax_{\Omega} \mathcal{L}(q,\Omega)$
\ENDWHILE
\ENDFOR
\end{algorithmic} 
\end{algorithm}
\subsection{Predictive Distribution}
Using the hyper-parameter $\Omega$ and the variational posterior distribution $q$, optimized by variational Bayesian learning, the predictive distribution of the $m$-th action $a_{*}^{(m)}$ and variance $\sigma_{*}^{2(m)}$ on an current state $\mathbf{s}_{*}$ can be computed analytically as (see \cite{rasmussen2003gaussian}).

\begin{figure*}[tb]
    \centering
    \includegraphics[width =0.9\hsize]{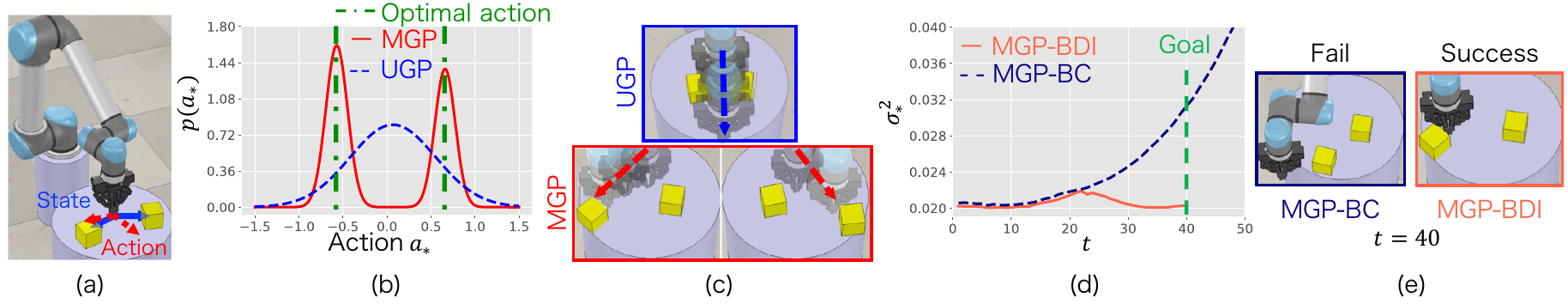}
    \caption{
    (a) V-REP environment used in the experiment of the table-sweep task for two boxes.
    (b), (c): Flexibility comparison between the MGP and UGP.
    (d), (e): Robustness comparison between the MGP-BDI and MGP-BC.
}
    \label{fig:V-REP}
\end{figure*}

\section{Evaluation}
In this section, the proposed approach is evaluated through simulation and experiments on a 6-DOF robotic manipulator. Specifically, to evaluate the robustness and flexibility of the proposed approach, the following key questions are investigated:
\romannumeral 1) Does the flexibility induced by MGP-BDI allow for capturing policies with multiple optimal behaviors?
\romannumeral 2) Does inducing model robustness via noise injection reduce covariate shift error?
\romannumeral 3) Is the optimized injection noise of significantly low variance, allowing for safe perturbations in real-world demonstrations?

To evaluate the proposed method (MGP-BDI), comparisons are made against baseline policy learning methods:
\begin{inparaenum}
\item Supervised imitation learning (\ie behavior cloning (BC)\cite{bain1995framework}) using standard unimodal GPs (UGP-BC) \cite{rasmussen2003gaussian},
\item BDI using standard unimodal GPs (UGP-BDI),
\item BC using infinite overlapping mixtures of Gaussian processes (MGP-BC).
\end{inparaenum}
Specifically, these three are chosen since they represent the state-of-the-art in either flexible or robustness imitation learning.
In all experiments, the maximum number of mixture GPs is fixed at ($M = 5$).
\begin{figure*}[t!]
    \centering
    \vspace{-2mm}
    \includegraphics[width =0.8\hsize]{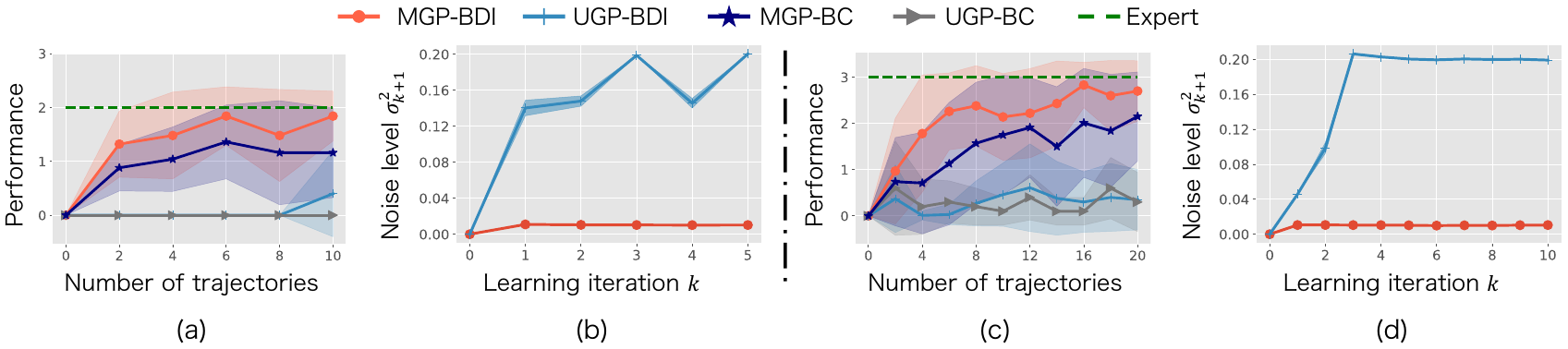}
    \caption{ 
    (a), (c): Comparing learning performance with the number of trajectories. The mean and standard deviation of the performances for 100 test trials.
    (b), (d): Comparing the tendencies of the injection noise parameters to be optimized according to the training iteration $k$.}
    \vspace{-5mm}
    \label{fig:Result}
\end{figure*}

\subsection{Simulation}
An initial experiment is presented to characterise the flexibility and robustness of the proposed approach, for learning and completing tasks in an environment. Specifically, a standard manipulator learning task (table-sweeping) involving multi-object handling is performed in the V-REP~\cite{rohmer2013VREP} environment, as shown in Fig.~\ref{fig:V-REP}-(a). 

In this experiment, learned policies are evaluated in terms of ability to flexibly learn tasks with multiple optimal actions (\eg the order in which to sweep the objects from the table), and well as robustness to environmental covariance shift inducing disturbances (\eg friction between the objects and table inducing variations of object movement). 

\subsubsection{Setup}
\label{s:sim_setup}
Initially, two boxes and the robot arm are placed at fixed coordinates on a table. The state of the system is defined as the relative coordinate between the robot arm and two boxes ($Q=4$), an action is defined as the velocity of the robot arm in the x and y axis. Demonstrations are generated using a custom PID controller to simulate the human expert, which sweeps the boxes away from the centre. Two demonstrations from these initial conditions are then performed, capturing both variations in the order of which the objects are swept from the table. 
For each demonstration, the performance is given by the total number of boxes swept off the table (min : 0, max: 2).
If a demonstration is unsuccessful (\ie both boxes were not swept off), the demonstration is discarded, and repeated. Given two successful demonstrations, the data is used to optimize the policy and noise parameters until (\ref{MGP:objective_function}) converges (as seen in Fig.~\ref{fig:overview}). This experiment is repeated for $K$ iterations, appending the successful demonstrations to the training dataset, and continuously updating the policy and noise estimates until convergence of the noise parameters. In the following experiments, $K=5$ results in convergence. During the test stage, only the robot arm is randomly placed at $\mathbf{s}_0\sim\mathcal{U}(\mathrm{centre}-0.01\mathrm{m},\mathrm{centre}+0.01\mathrm{m})$.

A second experiment is also presented, in a more complex task. To evaluate the proposed method's scalability, three boxes are placed in the environment, in random coordinates one by one within trisected areas on the table. 
The state is given by the relative coordinate between the robot arm and three boxes ($Q=6$), and the performance was calculated as the total number of boxes swept from the table (min : 0, max : 3). 
Due to the increased task complexity, the maximum number of iterations for updating the policy and noise estimates is $K=10$. 
\subsubsection{Result}
The results for these experiments are seen in Fig.~\ref{fig:V-REP}. In terms of flexibility of learning, Fig.~\ref{fig:V-REP}-(b) shows that the unimodal policy learner fails to capture the fact there are multiple optimal actions in the environment, instead learning a mean-centred policy that fails to reach either object (as seen in Fig.~\ref{fig:V-REP}-(c). In comparison, the proposed multi-modal approach correctly learns the multi-modal distribution, and outputs actions to sweep the two boxes. 

In terms of robustness, it is seen in Fig.~\ref{fig:V-REP}-(d) that even if a non-parametric flexible policy learner is used (MGP-BC), the variance in the learned action for this policy exponentially increases after the $20$-th time-step, and task failure occurs as seen in Fig.~\ref{fig:V-REP}-(e). This time-step indicates where the robot interacted with the object in the environment, and a sudden increase in uncertainty of the box position occurred. This is a possibility due to the dynamic behavior of the box being different that encountered between this demonstration and the training set, resulting in the model being unable to recover. In comparison, while the proposed noise-injected method also experiments some uncertainty at this interaction, it recovers and retains a near constant certainty throughout the remainder of the task application. 
In the two-box experiment, MGP-BDI and MGP-BC respectively earned performances $92\pm10.7\%$ and $58\pm34.7\%$ of the expert demonstrator, and in the three-box experiment this remained similar at $90\pm14.3\%$ and $71\pm30.9\%$. Other models in which unimodal policies are learned (UGP-BDI and UGP-BC) have been verified to yield about 0 performances.

To evaluate the learning performance, the performance and optimized noise variance is evaluated. In the two-box experiment, Fig.~\ref{fig:Result}-(a), it is seen that the unimodal policy methods (UGP-BC, and UGP-BDI), both fail to learn the multi-modal task, and retain a near zero performance throughout learning. In comparison, the multi-modal policy methods (MGP-BC, and MGP-BDI), both increase learning performance as the number of iterations increase, and MGP-BDI outperforms consistently. In addition to this, it is seen in Fig.~\ref{fig:Result}-(b) when the expert has probabilistic behavior (\ie multiple targets), the standard UGP-BDI learns an excessive (and potentially unsafe) amount of noise. This is due to the fact that because UGP-BDI assumes a deterministic policy model, a large modeling approximation error is induced, resulting in uncertainty in task performance, and overestimation in the noise variance when attempting to self-correct.

In contrast, the proposed method retains a very low variance in injected noise, by learning policy and noise optimization in a Bayesian framework allowing for confidence in the model to increase gradually as more training data sets are collected. This corresponds to a more stable, and safer demonstration. A similar set of results is seen in the more complex three-box experiment Fig.~\ref{fig:Result}-(c-d). 

\subsection{Real robot experiment with a human expert}
\begin{figure}
    \centering
    \includegraphics[width =0.8\hsize]{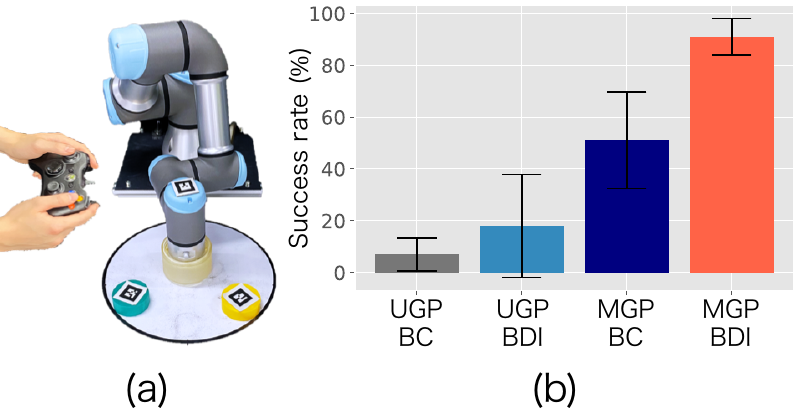}
    \caption{(a): Experimental environment for 6-DOF robotic arm (UR3) table-sweep task with human expert. 
    (b): Comparison of success rate. Success rate of each learning models is measured over 100 test trials.}
    \label{fig:Real}
\end{figure}
\subsubsection{Setup}
The proposed method is evaluated in a real-world robotics task, where the demonstrations of sweeping office supplies (\eg rubber tape) from a table in the correct order are  provided by a human expert, as shown in Fig.~\ref{fig:Real}-(a). 

Prior to the start of a demonstration, the two objects are placed at random positions in the upper semicircle of the table. Following the same procedure as outlined in Sec.\ref{s:sim_setup}, the human expert performs two demonstrations in which the objects are swept off the table in random order, and the method optimizes the measure and the noise parameter until (\ref{MGP:objective_function}) is converged. This process is repeated $K=5$ times (10 trajectories).
To measure the state of the system (the x,y position of the objects) AR markers are attached to each object (robot arm and rubber tape) and tracked through a RGB-D camera (RealSense D435). As in the previous simulation experiment, the task action is the end-effector velocity.
To validate the performance of MGP-BDI in reducing covariate shift, the initial positions of office supplies were arbitrarily placed in a circle on the table to induce scenarios this being a challenging task due to the covariate shift.
The learning models' performance is evaluated according to the success of the test episode.
Success is determined by the presence of office supplies on the table at the end of the test episode.

\subsubsection{Result}
The results of this experiment are seen in Fig.~\ref{fig:Real}-(b).
From this, it is seen that the unimodal policy methods (UGP-BC, UGP-BDI) have a poor success rate of $7\pm 6.4\%$ and $18\pm 19.9\%$, respectively. As such, they fail to correctly learn policies to account for multiple optimal actions in the environment, and demonstrate a lack of flexibility. In contrast, the multi-modal policy methods (MGP-BC and MGP-BDI) show improved performance ($51\pm 18.7\%$ and $91\pm 7\%$, respectively), however it is clear that even when incorporating flexibility, the success rate for BC is poor. 

\section{Conclusion}
This paper presents a novel Bayesian imitation learning framework, that injects noise into an expert's demonstration, to learn robust multi-optimal policies. This framework captures human probabilistic behavior and allows for learning reduced covariate shift policies, by collecting training data on an optimal set of states. The effectiveness of the proposed method is verified on a real robot with human demonstrations. In the future, this approach will be integrated with kernel approximation methods, or deep neural networks, to learn complex multi-action tasks from unstructured demonstrations  (\eg cooking involving cutting, mixing, and pouring).

\section{Appendix}{\label{Appendix}}

\subsection{Update laws in $q$}
Update of $q(\mathbf{f}^{(m)})$:
\begin{align}
q(\mathbf{f}^{(m)})&=\mathcal{N}(\mathbf{f}^{(m)} \mid \mu^{(m)}, \mathbf{C}^{(m)}) \label{Appendix:F}\\
\mu^{(m)} &=\mathbf{C}^{(m)} \mathbf{B}^{(m)} \mathbf{a}^* \\
\mathbf{C}^{(m)} &=(\mathbf{K}^{(m)^{-1}}+\mathbf{B}^{(m)})^{-1} \\
\mathbf{B}^{(m)} &=\operatorname{diag}\{r_{nm}/\boldsymbol\Sigma_{nn}\}
\end{align}

Update of $q(\mathbf{Z})$:
\begin{align}
q(\mathbf{Z}) &= \prod_{n=1}^{N} \prod_{m=1}^{\infty} r_{nm}^{\mathbf{Z}_{nm}}\label{Appendix:Z}, r_{nm}=\frac{\rho_{nm}}{\sum_{m=1}^{\infty} \rho_{nm}} \\
\log \rho_{nm}&={-\frac{1}{2}}\log{(2\pi\boldsymbol\Sigma_{nn})} - \frac{1}{2\boldsymbol\Sigma_{nn}}({a^*_n}-\mu_n^{(m)})^2+\psi(\alpha_{m}) \nonumber\\
-&\psi(\alpha_{m}+\beta_{m})+\sum_{j=1}^{m-1}\{\psi(\beta_{j})-\psi(\alpha_{j}+\beta_{j})\}
\end{align}
where, $\psi(\cdot)$ is the digamma function.

Update of $q(v_m)$:
\begin{align}
q(v_m) &= \operatorname{Beta}(v_{m} \mid \alpha_m, \beta_m) \label{Appendix:v}\\
\alpha_m &= 1+\sum_{n=1}^{N} r_{nm}, ~\beta_m = \beta+\sum_{j=m+1}^{\infty} \sum_{n=1}^{N} r_{nj} 
\end{align}

\subsection{Lower bound of marginal likelihood:}
\begin{align}
&\mathcal{L}(q,\Omega) \nonumber \\
&= \sum_{m=1}^{\infty} \log \mathcal{N}(\mathbf{a}^{*}\mid \mathbf 0,\mathbf{K}^{(m)}+\mathbf{B}^{(m)^{-1}})-\operatorname{KL}(q(\mathbf{v}) \mid\mid p(\mathbf{v})) \nonumber\\
&+\sum_{n=1}^{N}\sum_{m=1}^{\infty} r_{n m} \left\{ -\frac{1}{2} \log(2\pi \boldsymbol\Sigma_{nn}) + \psi(\alpha_m) - \psi(\alpha_m+\beta_m) \right. \nonumber \\
& \hspace{25mm}\left. + \sum_{j=1}^{m-1}\{\psi(\beta_{j})-\psi(\alpha_{j}+\beta_{j})\} - \log r_{nm} \right\} \nonumber\\
\end{align}

\bibliographystyle{IEEEtran}
\bibliography{ref}

\end{document}